%% file: main.tex
\documentclass[conference]{IEEEtran}
\usepackage{times}
\usepackage{multicol}
\usepackage{graphics,graphicx,caption,float,subcaption,booktabs,xcolor,multirow,array,color,ifthen,tabu,colortbl,dblfloatfix,url,xparse,mathtools,patchcmd,algorithm,algorithmic,amssymb,xspace,nicefrac,microtype,amsmath,amsfonts,bm,ragged2e,tikz,stackengine,etoolbox,xpatch,enumerate,xstring,setspace,tabularx,makecell,changepage,cuted,titlesec,wrapfig,sidecap,comment, bbding, placeins}
\usepackage{gensymb,comment}
\usepackage{listings}       %
\usepackage[most]{tcolorbox}   %
\definecolor{comment-green}{rgb}{0.435, 0.576, 0.106}
\definecolor{prompt-gray}{HTML}{a7a7a7}
\definecolor{light-gray}{rgb}{0.8, 0.8, 0.8}
\definecolor{light-yellow}{HTML}{f8f9cb}
\definecolor{light-blue}{HTML}{E6EEF9}
\definecolor{hyper-blue}{HTML}{367DBD}
\definecolor{comment-green}{HTML}{366B6B}
\definecolor{number-grey}{HTML}{656565}
\definecolor{string-red}{HTML}{B52020}
\definecolor{keyword-green}{HTML}{007E00}
\lstdefinestyle{code_lstlisting}{ %
    language=Python,                  
    backgroundcolor=\color{white},   
    basicstyle=\ttfamily\fontsize{8}{8}\selectfont,        
    breakatwhitespace=true,         
    breaklines=true,                 
    postbreak=\mbox{\textcolor{number-grey}{\tiny$\hookrightarrow$}\space}, 
    captionpos=b,                    
    commentstyle=\color{comment-green},    
    escapeinside={\%*}{*)},          
    keywordstyle=\bfseries\color{keyword-green},       
    stringstyle=\color{string-red},     
    numbers=none,                    
    numberstyle=\tiny\color{number-grey}, 
    stepnumber=1,                    
    numbersep=5pt,                   
    showspaces=false,                
    showstringspaces=false,          
    showtabs=false,                  
    tabsize=2,                       
}

\usepackage{hyperref}
\hypersetup{
colorlinks=true,
linkcolor=blue,
filecolor=magenta,      
citecolor=blue
}
\usepackage{siunitx}
\newcommand{\lzhang}[1]{\textcolor{black}{#1}}
\IEEEoverridecommandlockouts                              
\title{\LARGE \bf 
FUNCanon: Learning Pose-Aware Action Primitives via Functional Object Canonicalization for Generalizable Robotic Manipulation
}
\ifdefined\isanonymous
    \author{%
        Anonymous Authors
    \thanks{Affiliations withheld for double-blind review.
    }\\
    }
\else
    \author{
        Hongli Xu$^{2,3*}$, Lei Zhang$^{1,3*\dag}$, Xiaoyue Hu$^{2*}$,
        Boyang zhong$^{2}$,Kaixin Bai$^{1,3}$, \\ 
        Zoltán-Csaba Márton$^{3}$,
        Zhenshan Bing$^{2}$, Zhaopeng Chen$^{3}$, Alois Christian Knoll$^{2}$, Jianwei Zhang$^{1}$ 
        \thanks{* The first three authors contribute equally to this paper.}
        \thanks{\dag Corresponding author. zhanglei.cn.de@gmail.com, lei.zhang-1@studium.uni-hamburg.de}
        \thanks{$^{1}$TAMS (Technical Aspects of Multimodal Systems), Department of Informatics, University of Hamburg, Hamburg, Germany. 
        }
        \thanks{$^{2}$Technical University of Munich, Germany. 
        }
        \thanks{$^{3}$Agile Robots SE, Munich, Germany. 
        }
        \thanks{This work is supported by New Generation Artificial Intelligence-National Science and Technology Major Project (2025ZD0122903).}
    
    }
\fi

\begin{document}
    \makeatletter
    \let\@oldmaketitle\@maketitle
    \renewcommand{\@maketitle}{\@oldmaketitle
    \includegraphics[width=0.9\linewidth]{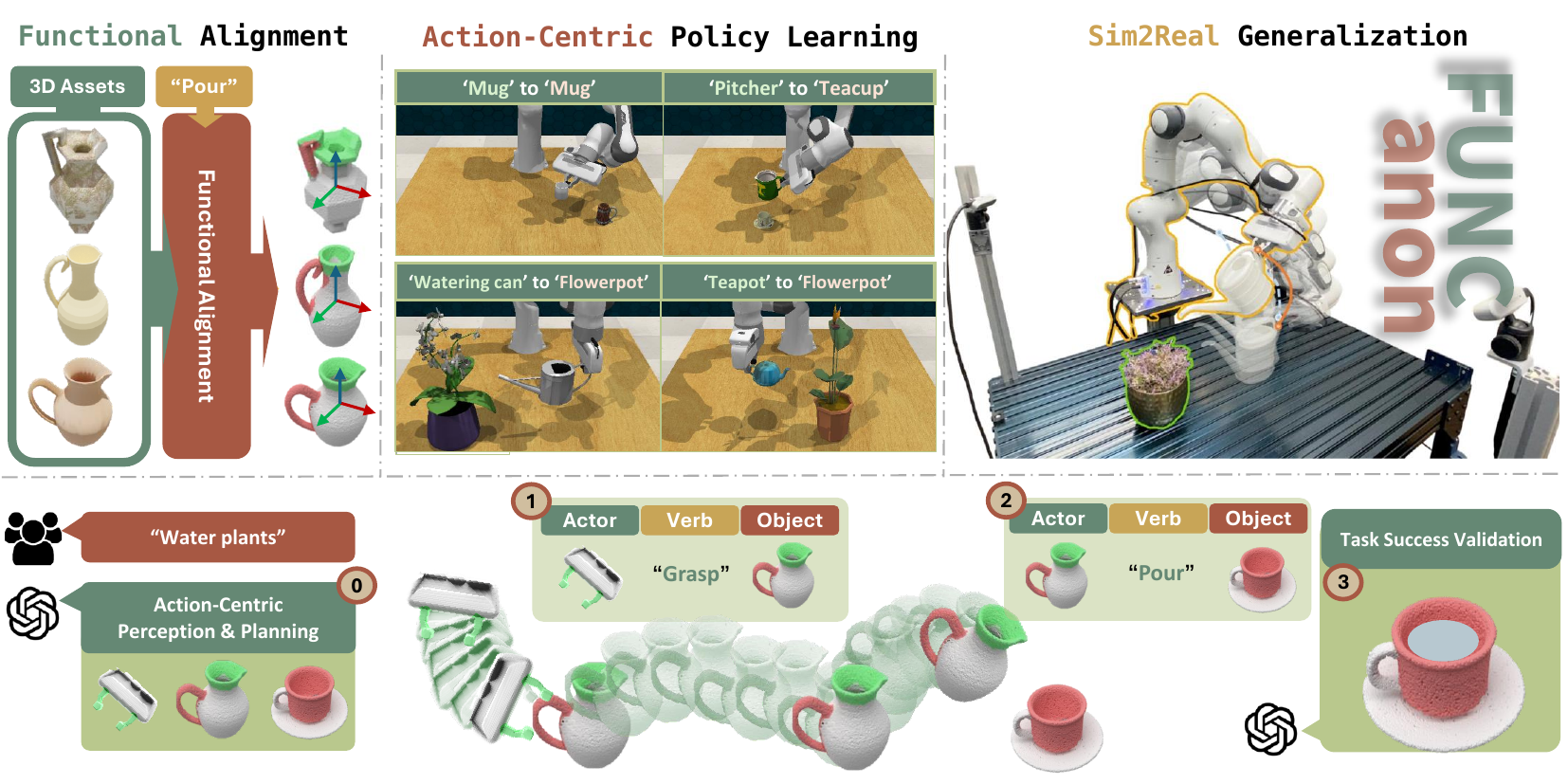}
      \centering
      \setcounter{figure}{0}
      \captionof{figure}{\small We present \textbf{FunCanon}, a framework that transforms long-horizon manipulation into action chunks defined by actor, verb, and object. Through \textbf{functional object canonicalization} using affordance cues from vision–language models, we \textbf{automatically generate manipulation trajectories} based on \textbf{functional alignment} and train an object-centric and action-centric diffusion policies \textbf{FuncDiffuser} that generalize across object instances, categories, and tasks, enabling action primitives for generalizable robotic manipulation and robust sim-to-real transfer.}
      \label{fig:teaser}
      \vspace{-0.1in}
      \bigskip}
\makeatother
\maketitle
\thispagestyle{empty}
\pagestyle{empty}

\setcounter{figure}{1}
\begin{abstract}
 General-purpose robotic skills from end-to-end demonstrations often leads to task-specific policies that fail to generalize beyond the training distribution. 
 Therefore, we introduce FunCanon, a framework that converts long-horizon manipulation tasks into sequences of action chunks, each defined by an actor, verb, and object. 
 These chunks focus policy learning on the actions themselves, rather than isolated tasks, enabling compositionality and reuse. To make policies pose-aware and category-general, we perform functional object canonicalization for functional alignment and automatic manipulation trajectory transfer, mapping objects into shared functional frames using affordance cues from large vision–language models. 
 An object-centric and action-centric diffusion policy FuncDiffuser trained on this aligned data naturally respects object affordances and poses, simplifying learning and improving generalization ability. 
 Experiments on simulated and real-world benchmarks demonstrate category-level generalization, cross-task behavior reuse, and robust sim-to-real deployment, showing that functional canonicalization provides a strong inductive bias for scalable imitation learning in complex manipulation domains. 
 Details of the demo and supplemental material are available on our project website~\url{https://sites.google.com/view/funcanon}.

\end{abstract}

\section{Introduction}
\input{sections/01_introduction.tex}

\section{Related Work}
\label{sec:related}

\input{sections/02_related_work.tex}

\section{Problem Statement and Methods}\label{sec:method}

\input{sections/03_methods.tex}

\section{Experiment}
\label{sec:experiment}
\input{sections/04_experiments.tex}

\section{Conclusion and Future Work}
\label{sec:conclusion}
\input{sections/05_conclusion.tex}

\vspace{0.1in}
{
\small
\bibliographystyle{IEEEtran}
\bibliography{main}
}
\input{sections/06_appendix.tex}
\clearpage
\FloatBarrier
\let\cleardoublepage\clearpage
\end{document}

%% file: sections/01_introduction.tex
As robots transition from controlled laboratory settings to unstructured real-world environments, developing robust and generalizable manipulation policies becomes increasingly critical. A fundamental challenge is enabling agents to generalize across unseen objects, diverse poses, and varying tasks — a capability that remains elusive for current imitation learning approaches. 

Imitation learning methods based on RGB images~\cite{chi2023diffusion} or point clouds~\cite{ze20243d} often suffer from limited precision and generalization due to viewpoint sensitivity, noisy observations, and redundant scene encodings. In contrast, 3D scene representations have shown promise in improving generalization~\cite{ke20243d}. To further address these challenges, object-centric representations, which focus on structured, object-level information such as 6D poses~\cite{hsu2024spot,wang20206} and scene flow~\cite{xu2020learning}, have gained significant attention. SPOT~\cite{hsu2024spot} demonstrates that SE(3) pose diffusion policy can improve cross-embodiment generalization, even when trained solely on passive human videos. 
Existing object-centric approaches~\cite{hsu2024spot} often depend on instance-specific, goal-conditioned trajectories, limiting generalization across categories and tasks. We attribute this to viewing manipulation as monolithic programs rather than modular, reusable behaviors. 
In the field of computer vision, prior work, such as UAD~\cite{tang2025uad} and Object Canonicalization~\cite{jin2025one}, mainly targets improving visual representations and semantic understanding. Related efforts have also investigated category-level affordance pose estimation~\cite{wang2024tooleenet}.  However, these methods have not explored how such representations can be leveraged for improving category-level alignment of robotic manipulation data, or augmenting manipulation trajectories.

Key open questions remain: \textit{how can generalized representations be leveraged to synthesize diverse manipulation data, model long-horizon tasks, and train robust, generalizable manipulation policies?}

To address these challenges, we propose \textbf{FunCanon}, a framework that models manipulation as compositions of reusable \emph{action primitives}—such as pouring, grasping, or inserting—defined over functionally aligned bi-object interactions. By leveraging affordance cues from large vision-language models~(VLMs), FunCanon canonicalizes semantically related objects (e.g., kettles and pitchers) into shared functional frames. This functional alignment enables the automatic manipulation trajectory transfer and the training of \emph{pose-aware, object-centric diffusion policies} that focus solely on interaction dynamics, decoupled from specific object identities, camera viewpoints, or task semantics. To achieve robust and generalizable manipulation, we first decompose long-horizon tasks into meaningful \emph{action chunks}, each specifying an actor, an action, and an object. This segmentation is performed by a large multimodal language model (MLLM), such as GPT-4o, in combination with large vision models (LVM) that extract affordance cues to guide chunking based on functional relevance. Next, we integrate these affordance cues with precise object pose estimates to perform \emph{functional alignment}, canonicalizing objects into shared functional frames. This process identifies action-related affordance regions and aligns bi-object poses, producing a semantically grounded representation of manipulation interactions. Leveraging this functional alignment, automatic trajectory transfer method is proposed to augment training data on RLBench base tasks to increase data diversity and functional coverage. During policy training and inference, our object-centric diffusion policy receives inputs encoding both affordances, poses of bi-object pairs, point clouds and action verb and estimate actions. 
Our main contributions are:

\begin{itemize}
    \item Introducing \textbf{FunCanon}, which decomposes complex manipulation tasks into reusable \emph{action primitives} grounded in functionally aligned object pairs. We explicitly incorporate the \emph{grasping phase} within action primitives, addressing a key gap in prior approaches.
    \item Leveraging large vision-language models for \textbf{affordance-driven functional canonicalization} for \textbf{functional alignment} and \textbf{automatic trajectory transfer}, enabling pose-aware and category-generalizable policy learning.
    \item Developing an object-centric diffusion policy trained on functionally aligned data, achieving both instance-level and category-level generalization and robust sim-to-real transfer.
\end{itemize}

%% file: sections/02_related_work.tex
\begin{figure*}[!htb]
    \centering
    \includegraphics[width=0.9\linewidth]{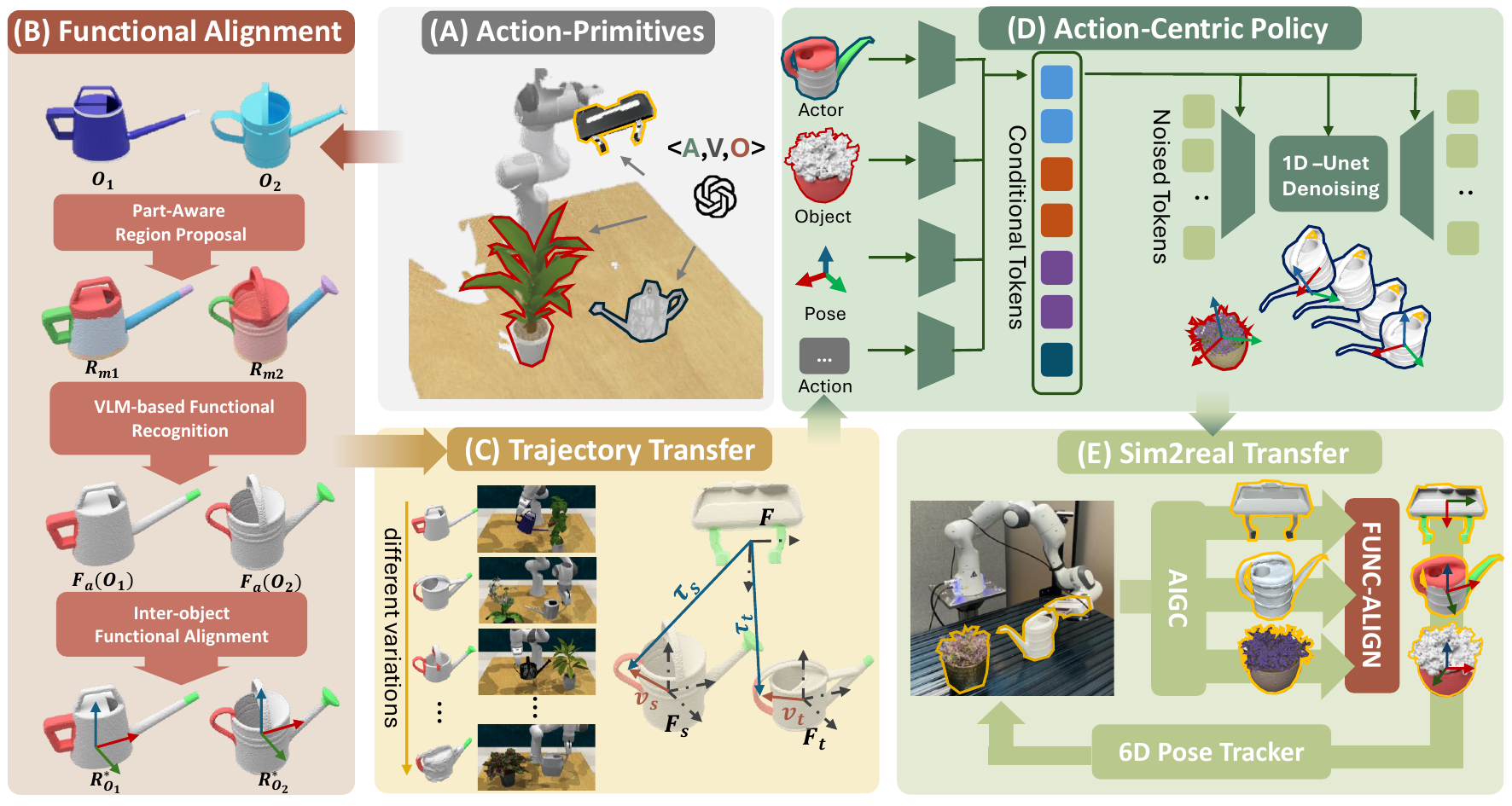}
    \caption{\small FunCanon Architecture. Our method consists of five components: (A) \textbf{Action Primitive Modeling}, where vision-language models perform AVO segmentation and task decomposition; (B) \textbf{Functional Alignment}, producing function-based canonicalization of objects; (C) \textbf{Automatic Trajectory Transfer}, which augments training data across instances and categories using online 3D datasets; (D) \textbf{Action-Centric Diffusion Policy learning} with FuncDiffuser, trained on this augmented data to predict AVO-centric actions; and (E) \textbf{Sim-to-Real Inference}, enabling robust transfer of the learned policy to real-world scenarios.}

    \label{fig.pipeline}
\end{figure*}

\subsection{Generalizable Robotic Manipulation}  
End-to-end visuomotor policies, trained via imitation~\cite{chi2023diffusion,ze20243d} or reinforcement learning~\cite{gu2017deep}, have laid the groundwork for robotic manipulation~\cite{zhang2025contactdexnet}. 
While effective on single, well-defined tasks, these approaches~\cite{chi2023diffusion,ze20243d} often face difficulties when scaling to long-horizon tasks, multimodal action distributions. To overcome these challenges, recent works incorporate structured priors~\cite{shridhar2023perceiver} or generative models~\cite{ke20243d} to improve robustness and generalization. For example, diffusion-based methods~\cite{ke20243d,mishra2023generative} have shown promising results in multi-task skill learning. The 3D Diffuser Actor (3DA)~\cite{ke20243d} further advances this line by conditioning policies on 3D scene representations, achieving better viewpoint robustness and sim-to-real transfer. 
Prior works typically achieve sim-to-real transfer through domain randomization or domain adaptation techniques to mitigate the domain gap between simulation and the real world~\cite{bai2024close,zhang2024collision,wang2025toward,bai2024cleardepth}. 
Despite these advances, these methods still tend to treat manipulation as monolithic, task-specific trajectory generation, limiting their flexibility and composability.

In contrast, object-centric representations focus on disentangling robot embodiment from task semantics by grounding policies in object frames. Techniques like CLIPort~\cite{shridhar2022cliport} and VIMA~\cite{jiang2023vima}
leverage language and vision to localize manipulation actions relative to object-centric coordinates. Category-level representations, such as Neural Descriptor Fields~\cite{simeonov2022neural} and keypoints~\cite{huang2024rekep}
, enhance generalization by capturing shared object properties. 
However, many object-centric approaches~\cite{hsu2024spot} remain tightly coupled to task-specific goals and trajectories, hindering reusability and broad generalization.

\subsection{Affordance Reasoning through LLM and VLM}  
Recently, VLMs and LLMs have been leveraged to reason about object affordances and functional parts~\cite{tang2025uad,chu20253d,zhu2025grounding,qian2024affordancellm} and plan functional robotic manipulation~\cite{li2024learning,zhang2025omnidexvlg}. 
UAD~\cite{tang2025uad}, distill affordance priors from foundation models to generalize without manual labeling. Geometry-semantic alignment methods like 
One-Shot 3D Canonicalization~\cite{jin2025one} further advance functional region understanding. While these advances demonstrate the potential of language-grounded affordance reasoning, most methods have yet to integrate these insights tightly with policy learning for generalizable robotic manipulation, increasing the instance-level and category-level generalization abilities.

%% file: sections/03_methods.tex
Robotic manipulation in unstructured environments requires policies that generalize across diverse objects, poses, and tasks.
Our goal is to learn generalizable manipulation policies that enable trajectory transfer across object instances without requiring additional human supervision. To this end, we propose an unified framework that integrates five key components: (A) Action primitive modeling and long-horizon task decomposition, 
(B) Functional alignment via vision-language and large language models, 
(C) Automatic and generalizable trajectory transfer, 
(D) Object-centric and action-centric diffusion policy learning, and 
(E) Sim-to-real inference for robust deployment on physical robots.

\subsection{Action Primitives and Task Decomposition}
We model manipulation as sequences of reusable action primitives. 
Formally, an action primitive is defined as a triplet $a = \langle \mathcal{A}, \mathcal{V}, \mathcal{O} \rangle$, 
where $\mathcal{A}$ represents the actor (e.g., kettle, pitcher), $\mathcal{V}$ denotes the action verb (e.g., grasp, pour), and $\mathcal{O}$ specifies the target object (e.g., cup, bowl). A long-horizon manipulation task can then be expressed as a sequence of action primitives $\mathcal{T} = \{a_1, a_2, ..., a_N\}$. 

We leverage large-scale multimodal language models $\mathcal{M}$ combined with semantic and functional cues extracted from vision models $\mathcal{V}$ to automatically perform task decomposition. 
This step not only focuses action learning on individual action semantics and object interactions but also facilitates subsequent action reuse and composition.

\begin{equation}
{a_1, a_2, ..., a_N} = \mathcal{D}(\mathcal{T}_e, \mathcal{M}, \mathcal{V})
\end{equation}
where $\mathcal{D}$ represents the decomposition function that takes task demonstrations $\mathcal{T}_e$ and models $\mathcal{M}$ and $\mathcal{V}$ as input, outputting action primitive sequences.

With this decomposition, our original goal of learning generalizable manipulation policies is reformulated into the problem of learning robust AVO-based action primitives. By treating each action primitive as a transferable motion segment, the challenge of generalization is shifted from entire trajectories to the representation and reuse of AVO triplets across objects and tasks.

\subsection{Functional Alignment}

Despite vast differences in appearance and category, different objects may assume similar operational functions. To achieve cross-category generalization of action primitives, we propose a functional alignment mechanism that unifies functional pose expressions across different objects. The key components contain of part-aware region proposal, VLM-based functional recognition and inter-object functional alignment. 

\subsubsection{Part-aware Region Proposal}
Given an object $o$ with its 3D model $M_o$, we first perform RGB-D rendering from multiple viewpoints to obtain corresponding RGB images and depth maps. Based on~\cite{tang2025uad}, we use a self-supervised visual feature encoder~\cite{oquab2023dinov2} to extract features from each view, and these 2D features are lifted to 3D space using the depth information, forming feature representations for candidate functional regions. We then cluster these features using KMeans to obtain $M$ functional region candidates, each represented as:
\begin{equation}
R_m = \{(x_i, f_i) \mid i \in C_m\},
\label{eq.functional_region_candidates}
\end{equation}
where $x_i$ denotes the 3D coordinates of a point, $f_i$ denotes the corresponding visual feature, and $C_m$ denotes the set of point indices belonging to the $m$-th region. Visualization of the feature space and clustering results are provided in the supplementary material.

\subsubsection{VLM-based Functional Recognition}
For each candidate functional region $R_m$, we extract its visual features $V_m$ and combine them with the object category information $c_o$. 
We then define a multimodal large language model as a \textit{binary classifier}:
\begin{equation}
\Phi(R_m, a, r, c_o) \to \{\text{True}, \text{False}\},
\end{equation}
where $a$ denotes an action (e.g., \textit{grasp}, \textit{pour}), and $r$ denotes a role hypothesis (active or passive). 
The function $\Phi$ evaluates whether region $R_m$ is functionally relevant to performing action $a$ under role $r$, conditioned on the object category $c_o$.  

The functional region set for object $o$ under action $a$ is then defined as:
\begin{equation}
F_a(o) = \{ (R_m, a, r) \mid \Phi(R_m, a, r, c_o) = \text{True} \}.
\end{equation}

This formulation explicitly links candidate regions to action--role pairs by verifying their functional relevance through binary decisions. 
For example, for a kettle, the handle region may be evaluated with the hypothesis $(\text{grasp}, \text{active})$, which yields \textit{True}, while the same region under $(\text{pour}, \text{passive})$ yields \textit{False}. 
As a result, the handle is explicitly associated with the grasp--active primitive but excluded from the pour--passive mapping.

\subsubsection{Inter-object Functional Alignment}

After obtaining the functional region set $F_a(o)$ for each
object, we leverage these regions to achieve functional
alignment between different objects, ensuring that action
primitives can be transferred reliably across object instances.
We assume that all object models have been normalized and
centered, so that each object is consistently scaled and
positioned at the origin.

For each object, we define a \emph{functional direction vector}
that captures the overall spatial location of its functional
regions:
\begin{equation}
v_o = \frac{1}{|F_a(o)|} \sum_{(x_i, f_i) \in F_a(o)} x_i,
\label{eq:vo}
\end{equation}
where $x_i$ denotes the 3D coordinate of a point in a functional
region. Intuitively, $v_o$ represents the ``functional center'' of
the object, serving as a reference for cross-object alignment.

Given a source object $o_s$ and a target object $o_t$, with
functional vectors $v_s$ and $v_t$, respectively, we aim to align
them by optimizing the object orientation. Specifically, the
alignment problem is formulated as:
\begin{equation}
R^* = \arg\min_{R \in SO(3)} \| R v_s - v_t \|_2^2 .
\label{eq:align-rot}
\end{equation}

Since most object models are aligned along the Z-axis, the only remaining rotational degree of freedom is the heading direction, i.e., rotation around Z. Therefore, finding the optimal rotation can be simplified to searching for a single angle around the Z-axis that best aligns the source and target vectors.

This formulation provides a lightweight yet effective strategy
for functionally consistent cross-object alignment, enabling
robust transfer of action primitives.

\subsection{Automatic and Generalizable Trajectory Transfer}
With functional alignment, we can leverage RLBench demonstrations for effective data augmentation. 
In RLBench, a complete robotic trajectory is typically composed of multiple sub-trajectories, 
where each sub-trajectory is generated by planning from a start pose to an end pose defined by relative constraints between a source object and a target object. 

Crucially, this decomposition naturally aligns with our AVO-based action primitives: each sub-trajectory corresponds to an action involving a specific actor, verb, and object. 
We exploit this alignment to augment training data as follows. 
Given a sub-trajectory $\tau_s$ associated with a source object $o_s$, we first transform the object coordinate system into its functional coordinate system defined by the functional vector $v_s$. 
Within this functional frame, the relative pose between the source and target objects is preserved, and the sub-trajectory is represented accordingly. 
To transfer the trajectory to a new object $o_t$, we simply apply the corresponding functional vector $v_t$ to map the trajectory back into the target object’s coordinate frame. 

Formally, the transferred trajectory $\tau_t$ is obtained as:
\begin{equation}
\tau_t = \tau_s + (v_t - v_s),
\end{equation}
where the addition denotes transforming the trajectory from the source functional frame to the target functional frame. 
By repeating this process across objects within a category, we can generate multiple variations of each action primitive, 
substantially increasing the diversity of training data while ensuring functional consistency. 
This enables the model to learn robust, generalizable AVO-based policies across different object instances and categories.

\subsection{Action-centric Policy Training}
We train a diffusion-based policy, named FuncDiffuser, to generate future object trajectories corresponding to action primitives, rather than directly predicting robot joint motions. 
Our state representation encodes the relative pose between actor and object as pose feature, the point clouds with associated functional areas as instance feature, and a CLIP-encoded embedding of the action verb to guide the motion. 
Formally, the state is defined as
\begin{equation}
s = (h_{\Delta}, f_A, f_O, v),
\end{equation}
where $h_{\Delta}$ denotes the \emph{relative pose feature} between the actor and object, $f_A, f_O$ are their visual functional features from PointNet++, and $v$ is the verb embedding. The policy $\pi_\theta(a \mid s)$ is trained using a diffusion-based denoising loss (MSE) to align predicted action trajectories with demonstration trajectories.

To enhance generalization, we leverage functional alignment and relative-pose features to perform \emph{trajectory augmentation}. RLBench trajectories are first mapped into their functional frames, and the relative poses are transferred across objects. 
This process generates multiple variations of each action primitive, enabling the diffusion policy to naturally respect object affordances and generalize across object instances and categories.

\subsection{Sim-to-Real Inference}
To enable robust deployment of our action-centric policies, we design a sim-to-real inference pipeline based on functional alignment. High-level human instructions are first processed by a large language model (GPT-4o), which decomposes the task into sub-tasks with actor--object pairs. Detailed decompositions examples are shown in appendix material. 

For each target object, a 3D mesh is generated using an AIGC model~\cite{TripoSR2024} to compensate for the lack of high-quality CAD models. Functional alignment is then applied to perform segmentation and alignment on the generated mesh, identifying regions relevant to the intended action.

Using these aligned functional regions, an off-the-shelf pose estimator~\cite{wen2024foundationpose} produces an object-centric scene representation, including the poses and spatial relations of functional regions. This representation, together with the planned actor--object pairs from task decomposition, is fed into the action-centric diffusion policy, which predicts trajectories for each sub-task. This pipeline enables accurate trajectory prediction and robust sim-to-r

%% file: sections/04_experiments.tex
\subsection{Experiments Setup}

\textbf{Simulation Environment:}~Experiments are conducted in RLBench~\cite{james2020rlbench}, a standard robotic manipulation benchmark providing diverse object categories, tasks, and demonstrations. This allows systematic evaluation of generalization under \textbf{pose-level}, \textbf{instance-level}, and \textbf{category-level} variations.

\begin{figure}[!tb]
    \centering\includegraphics[width=0.9\linewidth]{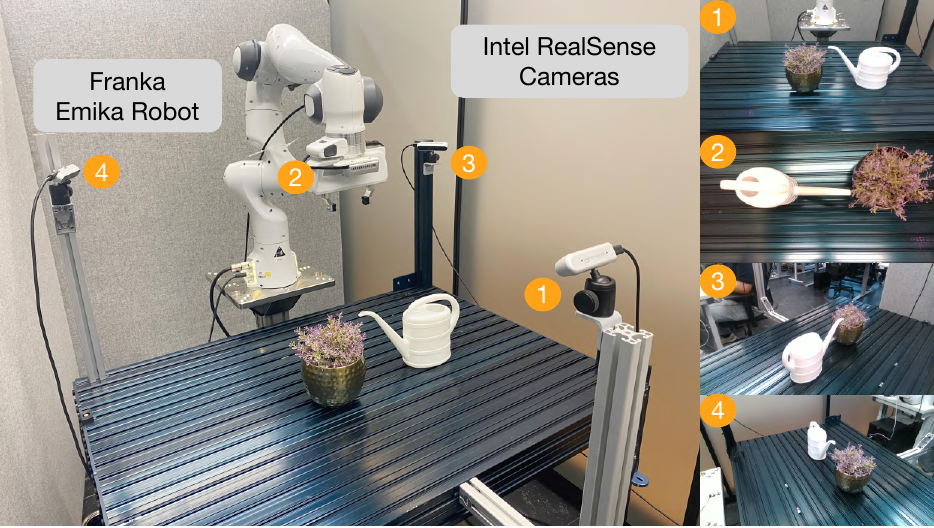}
    \caption{Real-World Experiment Setup with Franka Emika Robot and Intel RealSense Cameras.}
\label{fig.real_world_setup}
\end{figure}

\textbf{Real-world Environment:}~The real-world setup consists of a Franka Emika robot and multiple Intel RealSense cameras, as illustrated in Fig.~\ref{fig.real_world_setup}. This setup enables evaluation of sim-to-real transfer of our manipulation policies.

\subsection{Simulation Experiments: Data Generation}

We conduct all \textbf{pose-level}, \textbf{instance-level} and \textbf{category-level} manipulation experiments in a simulation environment based on RLBench~\cite{james2020rlbench}. During data collection, we monitor trajectory execution to annotate sub-task ground truth for each manipulation behavior, and record the corresponding \textbf{actor–manipulated object pairs}. We leverage \textbf{functional alignment} to automatically transfer trajectories between objects with similar functional roles. By aligning trajectories based on \textbf{functional alignment} of objects, our method generates new trajectory variants that preserve functional behavior rather than just low-level attributes like position or color.

\input{tabs/tab_instance_category_simulation_results_2.tex}

\textbf{Task and dataset Variation: }To systematically evaluate generalization ability, we consider three levels of task variation:

\textbf{Pose-level variation:} Using the original RLBench instances, tasks are executed with random initial poses of the objects and/or robot end-effector to test robustness to configuration changes.

\textbf{Instance-level variation:} objects within the same category are replaced with various shapes, sizes, or textures. All instance-level objects are sampled from Objaverse~\cite{deitke2023objaverse}.
\begin{figure}[!tb]
    \centering
    \includegraphics[width=\linewidth]{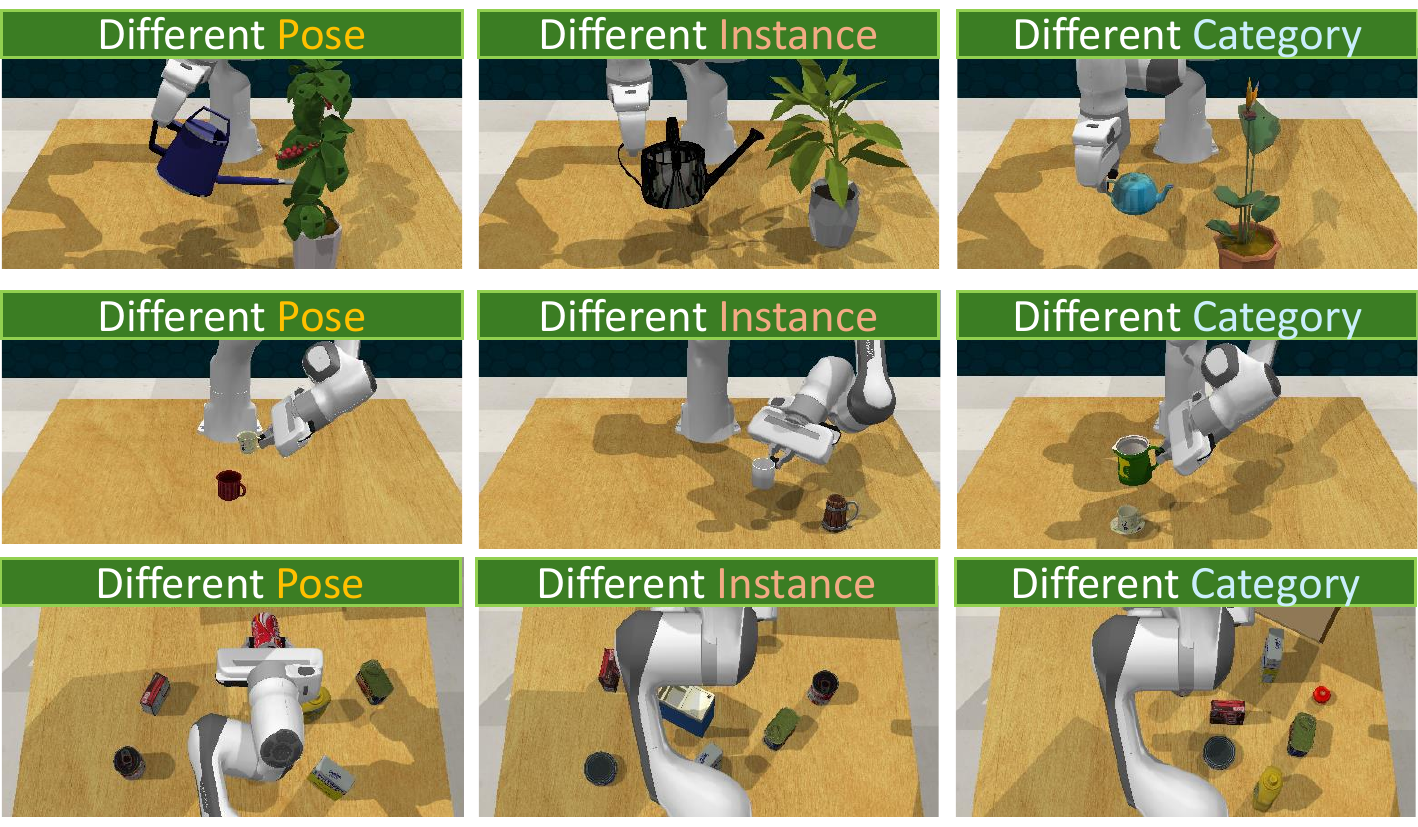}
    \caption{Task Variants in Simulation with Generated Trajectories based on our Functional Alignment method.}
    \label{fig.simulation}
\end{figure}

\begin{figure}[!ht]
    \centering
    \includegraphics[width=0.95\linewidth]{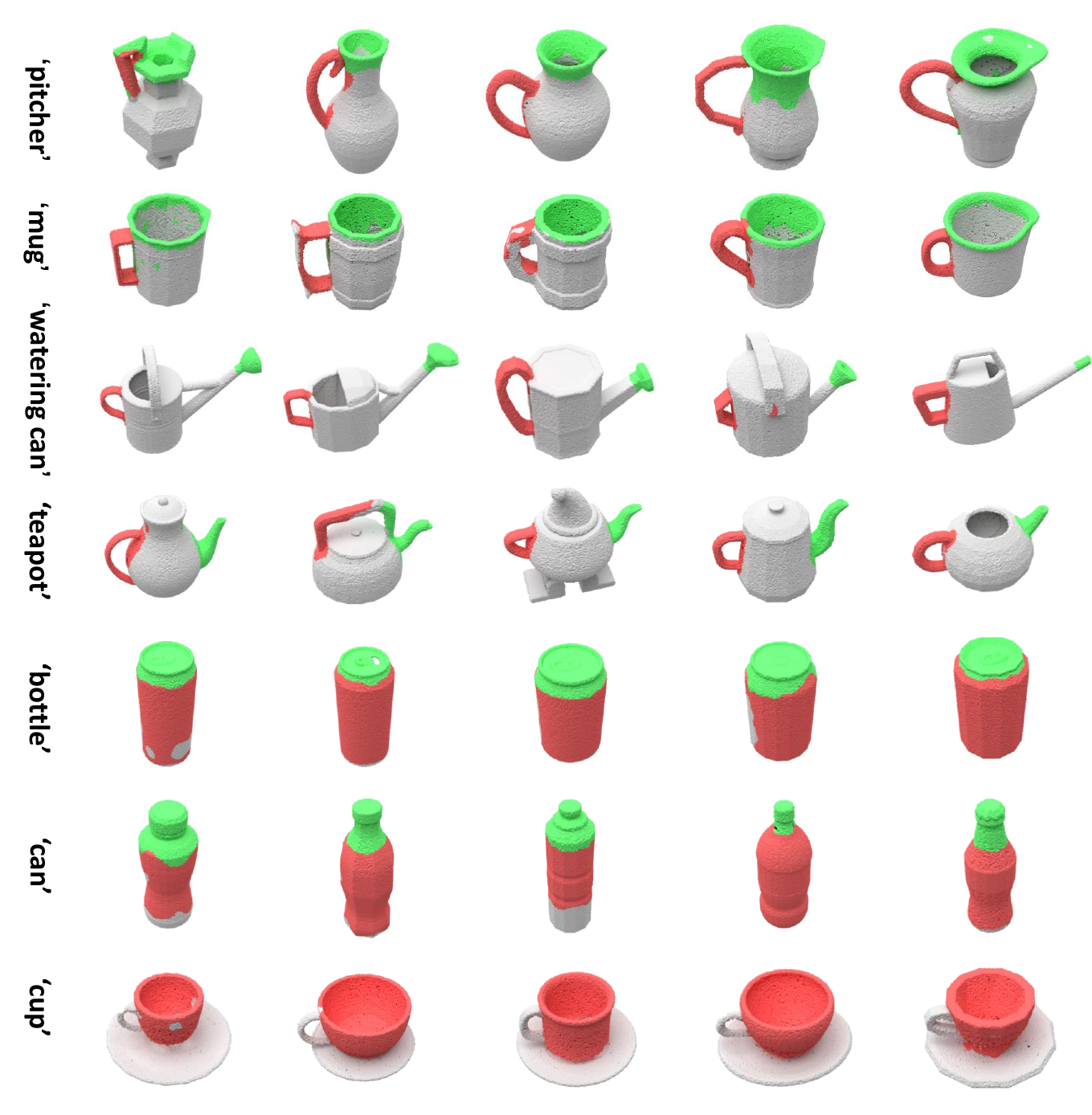}
    \caption{Qualitative Results of Functional Alignment.}
    \label{fig.functional_area_results}
\end{figure}

\textbf{Category-level variation:} objects are substituted with different but related categories or subcategories, introducing variations in geometry and functional parts (e.g., watering can vs. teapot). Fig.~\ref{fig.functional_area_results} shows examples of functional segmentation used to guide trajectory transfer and Fig.~\ref{fig.simulation} shows the generated trajectories for different variants.

\subsection{Implmentation Details}
The pipeline of real-world experiments consists of three main components: LLM-based task decomposition, pose estimation, and policy inference.
We employ GPT-4O for task decomposition, where the LLM takes human instructions and breaks down the overall task into sub-tasks with a list of actor-object pairs. 
We model the manipulation pipeline with multiple stages with different types of action primitives.
Then, we leverage FoundationPose~\cite{wen2024foundationpose} for pose estimation. 
For our diffusion policy network, the state encoder is a lightweight three-layer MLP that encodes the pose into a 64-dimensional representation. We additionally use a pretrained CLIP model to process the language instructions into sentence embeddings. The diffusion module follows a UNet-style backbone, optimized with AdamW at a learning rate of 1e-4. A DDIM scheduler with 100 timesteps during training and 10 at inference is employed. Training runs for 5000 epochs with a batch size of 64, ensuring stable convergence across tasks.
 
Based on the recognized poses and the planned actor-object pairs, the policy module infers the robot actions. Please refer to Supplementary Material for more implementation details. An example of our task decomposition method is shown as follows.

\input{code/task_decomposition_example.tex}

\subsection{Simulation Experiments: Policy Training Results}

We evaluate our policy's generalization through simulation experiments at both instance and category levels. For instance-level generalization, we train on a subset of objects within each category and validate on unseen instances from the same categories. For category-level generalization, training involves a few categories with representative instances, while validation includes unseen instances and novel categories to assess cross-category generalization.

\textbf{Evaluation Metrics:}~Success Rate (\textbf{SR}) measures the overall task completion, and Sub-task Success Rate (\textbf{Sub-task SR}) evaluates the completion of individual steps within each task.

\textbf{Tasks:}~We evaluate our policy on three core manipulation tasks: Put A in B, Pour A in B, and Water B with A. \lzhang{For trajectory transfer, we adopt the corresponding RLBench demonstrations from Put Groceries in Cupboard, Pour from Cup to Cup, and Water Plants. Specifically, Put A in B maps to the Pick \& Place task, Pour A in B to Pick, Pour (Level 1), and Water B with A to Pick, Pour (Level 2). The key difference between the two pouring tasks is the container design: Pour A in B uses a generic cup without a spout, while Water B with A employs a watering can whose spout directs the liquid flow.} We compare our method with several SOTA baseline approaches, including SPOT~\cite{hsu2024spot}, 3DA~\cite{ke20243d}, under all pose-level instance-level and category-level variations.

\textbf{Results and Analysis}: The results are summarized in Tab.~\ref{tab:sim_results}. The results demonstrate that our policy achieves high success rates across tasks and object variations, consistently outperforming existing methods, and effectively generalizing to unseen instances and categories. 

Our method outperforms SPOT and 3DA for several reasons. First, all methods share the same base instance, ensuring a fair comparison. Second, both our method and SPOT leverage pose-aware object policies, which gives them an advantage over 3DA. However, our evaluation is more challenging than SPOT, as we assess the complete manipulation pipeline starting from grasping, whereas SPOT begins from the post-grasp stage. Notably, the gap between instance-level and category-level success rates is larger for SPOT and 3DA, whereas our approach maintains high performance. This improvement is largely due to our functional alignment module, which enhances generalization to unseen object instances and categories. In contrast, SPOT relies on per-instance tracking and struggles to generalize to objects with novel coordinate conventions.

\input{tabs/tab_ablation_study_performance.tex}

\subsection{Ablation Study}

To analyze the contribution of key components in our policy, we perform an ablation study focusing on three main variants: \textbf{(A) Monolithic Trajectory}, \textbf{(B) No Functional Frame}, and \textbf{(C) Geometry-only}. Monolithic Trajectory removes multi-stage modeling by merging all stages into a single trajectory. No Functional Frame uses only world or camera coordinates, omitting functional keypoints. Geometry-only relies solely on geometric information, ignoring affordance and functional cues.

The results show that each component plays a crucial role in policy performance. Monolithic Trajectory performs poorly on complex multi-stage tasks, demonstrating the necessity of stage modeling. No Functional Frame reduces the success rate of grasping and manipulation steps, highlighting the importance of functional keypoints for pose-aware execution. Geometry-only also lowers generalization to unseen instances and categories, confirming the value of affordance and functional information. Overall, these findings validate that stage modeling, functional frames, and functional cues are essential for achieving high Success Rate (SR) and Sub-task Success Rate (Sub-task SR).

\begin{figure}[!htb]
    \centering
    \includegraphics[width=0.9\linewidth]{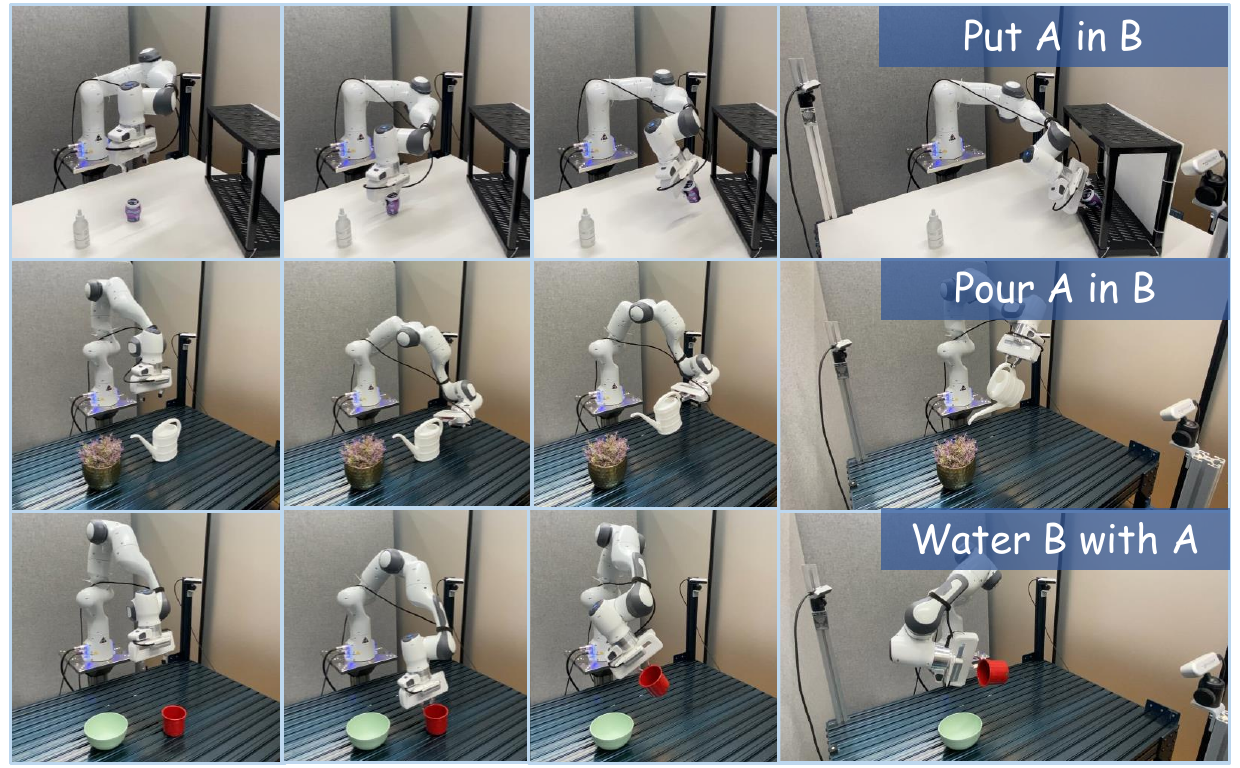}
    \caption{\small Qualitative Results of Real-World Experiments.}
    \label{fig.results_semantic_alignment}
\end{figure}

\subsection{Real-World Experiments (SIM2REAL Transfer)}

To validate the generalization of our policy, we conduct SIM2REAL transfer experiments in the real world. The policy is trained entirely in simulation and deployed directly on the real robot without any fine-tuning. 
We conducted 50 real-world trials with unknown objects, as shown in Fig.~\ref{sec:experiment}. The success rates are reported in Tab.~\ref{table.grasping_success_rate_real_world}. Our method consistently outperforms the baselines. Compared with SPOT, FunCanon improves success by at lease +14\%. 
The advantage is even more pronounced over 3DA, with improvements of at least +26\%.
Notably, the performance gap is largest in the pouring tasks, where functional alignment provides a strong inductive bias for handling different container geometries. 
These results validate that FunCanon generalizes more robustly across object variations and achieves reliable sim-to-real transfer.
\input{tabs/tab_SR_real_world.tex}

%% file: tabs/tab_instance_category_simulation_results_2.tex
\begin{table*}[!t]
\centering
\small
\caption{Success rate (\%) in simulation under pose, instance-level, and category-level substitutions across different methods (25 trials, three repetitions). Numbers indicate mean $\pm$ std. Avg row shows improvement over 3DA in green. Evaluation of 25 episodes using 3 random seeds for each task.}
\label{tab:sim_results}
  \setlength{\tabcolsep}{1.15mm}
  \resizebox{0.8\textwidth}{!}{
\begin{tabular}{lccccccccc}
\toprule
\textbf{Task} &
\multicolumn{3}{c}{\textbf{Ours (Funcanon)}} &
\multicolumn{3}{c}{\textbf{SPOT}} &
\multicolumn{3}{c}{\textbf{3DA}} \\
\cmidrule(r){2-4} \cmidrule(r){5-7} \cmidrule(r){8-10}
& Pose & Inst. & Cat. 
& Pose & Inst. & Cat. 
& Pose & Inst. & Cat. \\
\midrule
T1 & 60.0$\pm$3.27 & 61.3$\pm$3.77 & 60.0$\pm$3.27 & 
48.0$\pm$3.27 & 50.7$\pm$4.99 & 48.0$\pm$3.27 & 
28.0$\pm$3.27 & 33.3$\pm$1.89 & 32.0$\pm$3.27 \\
T2 & 77.3$\pm$4.99 & 80.0$\pm$3.27 & 72.0$\pm$3.27 & 
68.0$\pm$3.27 & 64.0$\pm$3.26 & 70.7$\pm$1.89 & 
24.0$\pm$3.27 & 25.3$\pm$1.89 & 22.7$\pm$1.89 \\
T3 & 80.0$\pm$3.27 & 84.0$\pm$3.27 & 72.0$\pm$3.27 & 
68.0$\pm$3.27 & 72.0$\pm$3.27 & 64.0$\pm$3.27 & 
36.0$\pm$3.27 & 40.0$\pm$3.27 & 32.0$\pm$3.27 \\
\midrule
\textbf{Avg.} 
& \textbf{72.4} \textcolor{green!60!black}{(+43.1)} 
& \textbf{75.1} \textcolor{green!60!black}{(+42.2)} 
& \textbf{68.0} \textcolor{green!60!black}{(+39.1)} 
& 61.3 \textcolor{green!60!black}{(+32.0)} 
& 62.2 \textcolor{green!60!black}{(+29.3)} 
& 60.9 \textcolor{green!60!black}{(+32.0)} 
& 29.3 & 32.9 & 28.9 \\
\bottomrule
\end{tabular}
}
\end{table*}

%% file: code/task_decomposition_example.tex
\begin{tcolorbox}[boxsep=0pt,
left=3pt,
right=3pt,
top=3pt,
bottom=3pt,
arc=0pt,
boxrule=0.5pt,
colframe=light-gray,
colback=white,
breakable,
enhanced]
  \begin{minipage}{1.0\textwidth}
  \textbf{Context: }\\\textit{\small Consider you are a 6-DOF robotic arm.
I will give you a daily-life task.
Decompose this task into a sequence of \textbf{core, action-centered steps}.\\
\textbf{Examples:}
{\small
\{
  "task": "pour water",
  "steps": [
    \{
      "step": 1,
      "action": "grasp",
      "actor": "robot gripper",
      "object": "teapot"
    \},
    \{
      "step": 2,
      "action": "pour",
      "actor": "teapot",
      "object": "cup"
    }
  ]
\}
\}\\}
\textbf{Output Format: }
\begin{lstlisting}[style=code_lstlisting]
{"task": "<task description>",
  "steps": [{
      "step": 1,
      "action": "<primitive action>",
      "actor": "<agent performing the action, e.g., robot gripper or tool>",
      "object": "<object being manipulated or affected>"
    },
    ...]}
\end{lstlisting}
\textbf{Instruction: }\textit{\small Stuck red cup into blue cup, then stuck green cup into blue cup.
}
  \end{minipage}
\tcblower
\textbf{Results: }
\small
\begin{lstlisting}[style=code_lstlisting]
{"task": 
    "stuck red cup into blue cup then stuck green cup into blue cup",
"steps": [
    {"step": 1,
    "action": "grasp",
    "actor": "robot gripper",
    "object": "red cup"},
    {"step": 2,
    "action": "insert",
    "actor": "red cup",
    "object": "blue cup"}
    ...]
}
\end{lstlisting}
\end{tcolorbox}

%% file: tabs/tab_ablation_study_performance.tex
\begin{table}[!t]
\centering
\small
\caption{\small Policy training of Pick, Pour (L2) under category-level substitutions (25 trials, one variation). 
SR: Success Rate; Sub-SR: success rate of each sub-task. Sub2-SR = Sub2 success / Sub1 success. Numbers in parentheses indicate performance drop relative to Ours (FunCanon).}
\label{tab:pick_pour_cat}
  \setlength{\tabcolsep}{1.15mm}
  \resizebox{0.9\linewidth}{!}{
\begin{tabular}{lccc}
\toprule
Method & Sub1-SR (\%) & Sub2-SR (\%) & SR (\%) \\
\midrule
Ours (FunCanon)   & 64.0 & 62.5 & 40.0 \\
Geom-only          & 60.0 & 60.0 & 36.0 \textcolor{red!70!black}{(-4.0)} \\
No-Func-Frame      & 44.0 & 54.5 & 24.0 \textcolor{red!70!black}{(-16.0)} \\
Monolithic         & 36.0 & 44.4 & 16.0 \textcolor{red!70!black}{(-24.0)} \\
\bottomrule
\end{tabular}
}
\end{table}

%% file: tabs/tab_SR_real_world.tex
\begin{table}[t!]
  \centering
    \caption{\small Real World Experiment Results. 
    }
\label{table.grasping_success_rate_real_world}
  \resizebox{1.0\linewidth}{!}{
    \setlength{\tabcolsep}{3.pt}
    \begin{tabular}{lcccc}
      \toprule
        \textbf{Methods} & \textbf{Pick, Place} & \textbf{Pick, Pour (L1)} & \textbf{Pick, Pour (L2)} \\
      \hline
  3DA~\cite{ke20243d} & 48\% & 54\% & 62\% \\
SPOT~\cite{hsu2024spot} & 52\% & 76\% & 60\% \\
\rowcolor[gray]{0.8}
\textbf{FunCanon (Ours)} & 88\% & 90\% & 88\%\\
      \midrule
    \end{tabular}
    }
  \vspace{3pt}    

\end{table}

%% file: sections/05_conclusion.tex
In this work, we introduced \textbf{FunCanon}, a framework that bridges object-centric and action-centric learning via \textbf{functional object canonicalization}. By converting long-horizon manipulation into modular \textit{actor–verb–object primitives} and aligning objects into functional frames, our method supports \textbf{automatic trajectory transfer} and policy learning that generalizes across objects, categories, and tasks. Simulation and real-world experiments validate that FunCanon substantially outperforms SOTA baselines and enables robust sim-to-real transfer. 
In simulation, our success rate improves by \lzhang{+14.6\%} over SPOT and \lzhang{+38.4\%} over 3DA. 
In real-world trials, FunCanon consistently outperforms baselines, surpassing SPOT by at least \lzhang{+14\%} and 3DA by at least \lzhang{+26\%} across all tasks.
The largest improvements appear in pouring tasks, where functional alignment enables robust handling of container geometries and spout orientations. These results highlight that FunCanon not only achieves higher success rates but also delivers stronger instance- and category-level generalization through functional canonicalization.

In future work, we plan to extend FunCanon to articulated object manipulation, and explore scalable data generation pipelines that further enhance cross-domain generalization.

%% file: sections/06_appendix.tex
\newpage
\appendix
Supplementary Materials consist of the following sections:
\begin{itemize}
    \item Instance counts and categories of objects for simulation experiments, as detailed in Sec.~\ref{appendix.instance_count}.
    \item Implementation details of sub-task decomposition in Sec.~\ref{appendix.sub-task_decomposition}
    \item Implementation details of simulation experiments in Sec.~\ref{appendix.details_sim_exp}.
    \item Implementation details of real-world experiments in Sec.~\ref{appendix.details_real_exp}.
    \item Analysis of Failures in Sec.~\ref{appendix.error_analysis}.
    \item Discussion of limitation in Sec.~\ref{appendix.limitations}.
    \item Additional results of functional alignment in Sec.~\ref{appendix.sec.additional_results_functional_alignment}.
    \item Additional results of automatic and generalizable trajectory transfer in Sec.~\ref{appendix.sec.additional_results_traj_transfer}.
\end{itemize}

\subsection{Instance Count with Category Information}
\label{appendix.instance_count}
Table~\ref{tab:instance_stats} summarizes the number of instances used for each category, along with their corresponding roles (e.g., Actor or Object) in different manipulation tasks.

\subsection{Implementation Details of Sub-Task Decomposition}
\label{appendix.sub-task_decomposition}
In our RLBench experiments, high-level instructions are decomposed into executable sub-tasks using a large language model (GPT-4O) as a task planner. The LLM receives a natural language description of a daily-life manipulation task and generates a sequential plan of primitive actions. Each sub-task specifies the \textit{actor performing the action} and the \textit{object being manipulated}, grounding every step in concrete object-centric interactions. 
This approach structures the plan using spatial constraints and interaction primitives.

During inference in RLBench, the LLM produces a series of ordered sub-tasks, such as grasping, lifting, inserting, or pouring objects. These sub-tasks are directly interpretable by the policy module, enabling the robot to execute them sequentially while reasoning about the states and poses of relevant objects. By grounding each step in actor-object pairs, the decomposition provides a \textit{structured and modular interface} between high-level language instructions and low-level robot actions, facilitating generalization across tasks with varying objects and configurations.

For example, a task such as ``pour water from a teapot into a cup'' in RLBench would be decomposed into object-centric manipulation steps, each specifying which object to grasp, how to manipulate it, and which target object is affected. This ensures that the resulting plan can be smoothly executed by the downstream policy network.
\input{tabs/tab_instance_count_with_category.tex}

\subsection{Implementation Details of Simulation Robotic Experiments}
\label{appendix.details_sim_exp}

\section{Implementation Details of Simulation Robotic Experiments}

In our simulation experiments on RLBench, the inference process executes high-level manipulation tasks in a fully structured, sub-task-wise manner while maintaining an \textit{action-in-the-loop} paradigm. Given a natural language instruction, such as ``pour water from the teapot into the cup'' or ``insert red cup into blue cup,'' the instruction is first processed by GPT-4O, which decomposes it into a sequence of object-centric primitive actions. Each sub-task specifies the actor performing the action and the target object, grounding every step in concrete object interactions.

\begin{figure*}[t]
    \centering
    \includegraphics[width=0.95\textwidth]{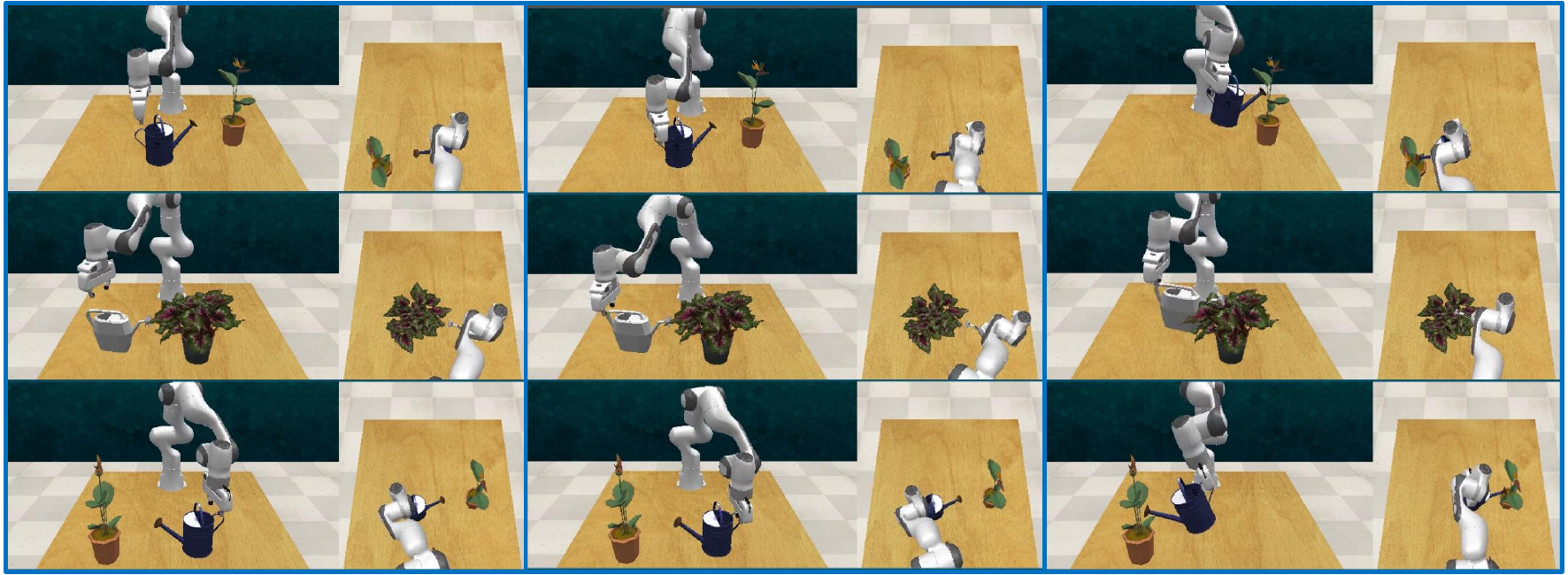}
    \caption{Visualization of inference results in RLBench simulation. Each snapshot shows the robot executing sub-tasks generated by the LLM, conditioned on the current 6D object poses.}
    \label{fig:inference_results}
\end{figure*}

To provide accurate state information for the policy network, we track the 6-DoF poses of all relevant objects at each timestep. In simulation, poses are obtained via a 6D pose tracking module ~\cite{wen2024foundationpose}. The diffusion-based policy network operates in a closed loop with the environment: at each timestep, it predicts robot actions conditioned on the current object poses while following the sub-task instructions from the LLM. This design allows the policy to react dynamically to changes in object positions and orientations, enabling robust execution even under perturbations. Each predicted action is executed sequentially, and the resulting state (updated 6D object poses) is fed back to the policy network for the next step. This \textit{action-in-the-loop} setup, combined with continuous 6D pose tracking, ensures that the robot can accurately adjust its actions based on the latest object configurations. Tasks are considered successful if the final object configuration matches the desired outcome. 

we provide a visual summary of the inference process to demonstrate the effectiveness of our approach. Figure~\ref{fig:inference_results} shows a sequence of snapshots taken during the execution of a high-level manipulation task. The LLM provides sub-task instructions, which are interpreted by the diffusion-based policy network conditioned on the current 6-DoF poses of objects. Each frame highlights the actor-object interactions corresponding to a primitive action, such as grasping, lifting, pouring, or inserting objects.

This visualization emphasizes the \textit{action-in-the-loop} nature of our method: at every timestep, the robot observes the updated object poses, predicts the next action, and executes it, creating a closed feedback loop between perception and action. By following the LLM-provided sub-task plan while continuously tracking object poses, the robot can adapt to changes in object positions and orientations, ensuring accurate task completion.

The inference results in Figure~\ref{fig:inference_results} illustrate not only the successful completion of tasks but also the smooth transition between sub-tasks, confirming that the system effectively integrates language-guided decomposition, 6D pose tracking, and diffusion-based policy execution.

\subsection{Implementation Details of Real-World Robotic Experiments}
\label{appendix.details_real_exp}

\begin{figure}[hb]
    \centering
    \includegraphics[width=0.9\linewidth]{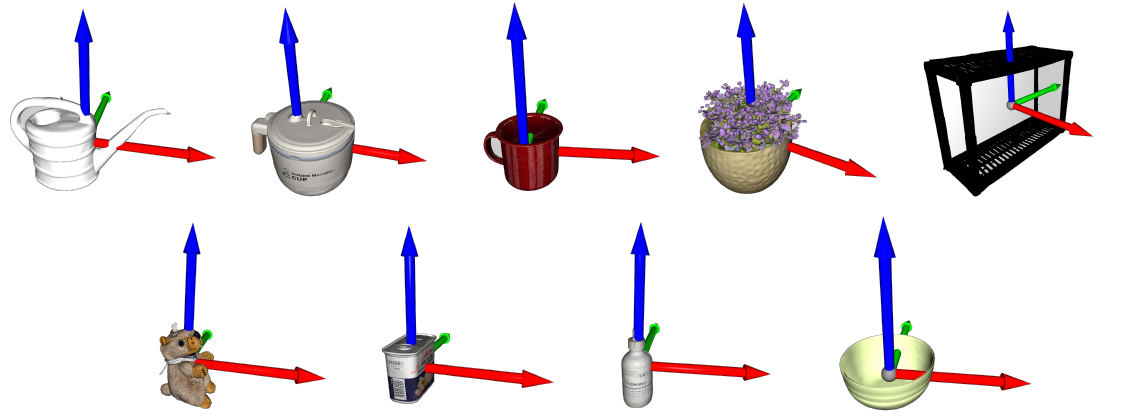}
    \caption{Object Reconstruction Results, including watering can, instant noodle cup, mug, flower, shelf, dolomiti bear, lunch meat, dropper bottle, bowl.}    \label{fig.appendix.object_reconstruction_result}
\end{figure}

The reconstructed objects with textures are shown in Fig.~\ref{fig.appendix.object_reconstruction_result}. 

\subsection{Analysis of Failures}
\label{appendix.error_analysis}
The primary causes of failure are as follows. Firstly, the pose estimation algorithm exhibits instability when dealing with symmetric objects or those lacking distinctive semantic features. Furthermore, when the object is in motion, the accuracy of pose updates becomes insufficient, which can result in policy execution failure. This issue is also partly attributed to the quality of sensor data. Specifically, the limitations in accuracy and noise present in RealSense data can significantly affect overall system performance. Occluded scenes further increase the difficulty of pose estimation. 
Errors in the pose tracker have a greater impact on functionally constrained tasks, such as pouring. 
Secondly, the presence of complex shapes or handles on smaller cups could also increase the difficulty of manipulation tasks. 
Lastly, execution failures also occur due to target poses being located outside the robot's reachable workspace, resulting in kinematic inaccessibility. To enable a fairer evaluation of Sim2Real performance, such cases are excluded.

\begin{figure}[!t]
    \centering
    \includegraphics[width=0.85\linewidth]{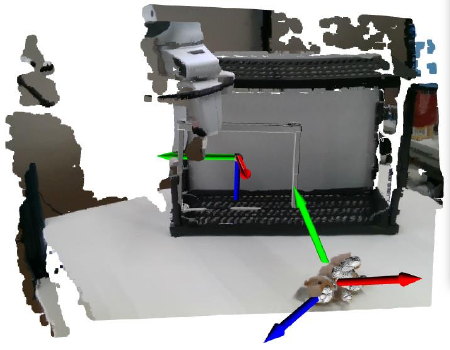}
    \caption{Example of failed pose tracking. As shown in the figure, the shelf often leads to failure due to its symmetry and lack of texture.}
    \label{fig.appendix.failure_object_tracking}
\end{figure}

\subsection{Limitations}
\label{appendix.limitations}

Although FunCanon achieves strong generalization across simulated and real-world tasks, it faces several limitations. The simulation environment lacks support for critical behaviors such as re-grasping, and cannot execute actions outside the predefined behavior set, limiting task diversity and complexity. Additionally, the framework relies heavily on accurate 6D object pose estimation, which may fail in real-world conditions involving occlusions, symmetry, or sensor noise—particularly in pose-sensitive tasks like pouring.

\subsection{Additional Results of Functional Alignment.}
\label{appendix.sec.additional_results_functional_alignment}
\begin{figure*}[!thb]
    \centering
    \includegraphics[width=0.85\linewidth]{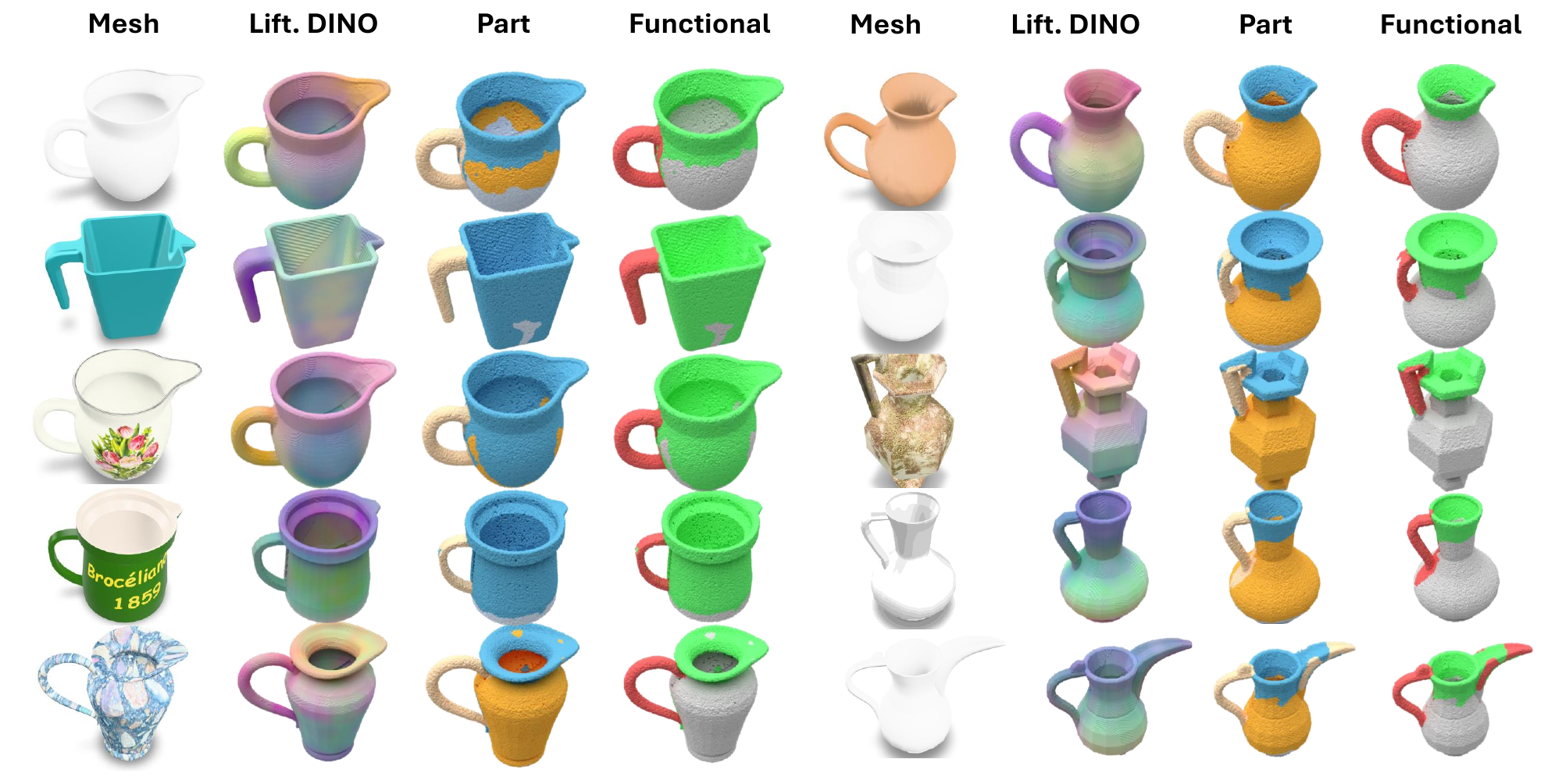}
    \caption{Additional results of functional alignment, from 3D self-supervised feature visualization, part-aware region proposal, functional alignment results.}
    \label{fig.appendix.results_functional_alignment}
\end{figure*}

Additional results of functional alignment are shown in Fig.~\ref{fig.appendix.results_functional_alignment}.

\subsection{Additional Results of Automatic and Generalizable Trajectory Transfer}
\label{appendix.sec.additional_results_traj_transfer}
\begin{figure*}[!thb]
    \centering
    \includegraphics[width=0.8\linewidth]{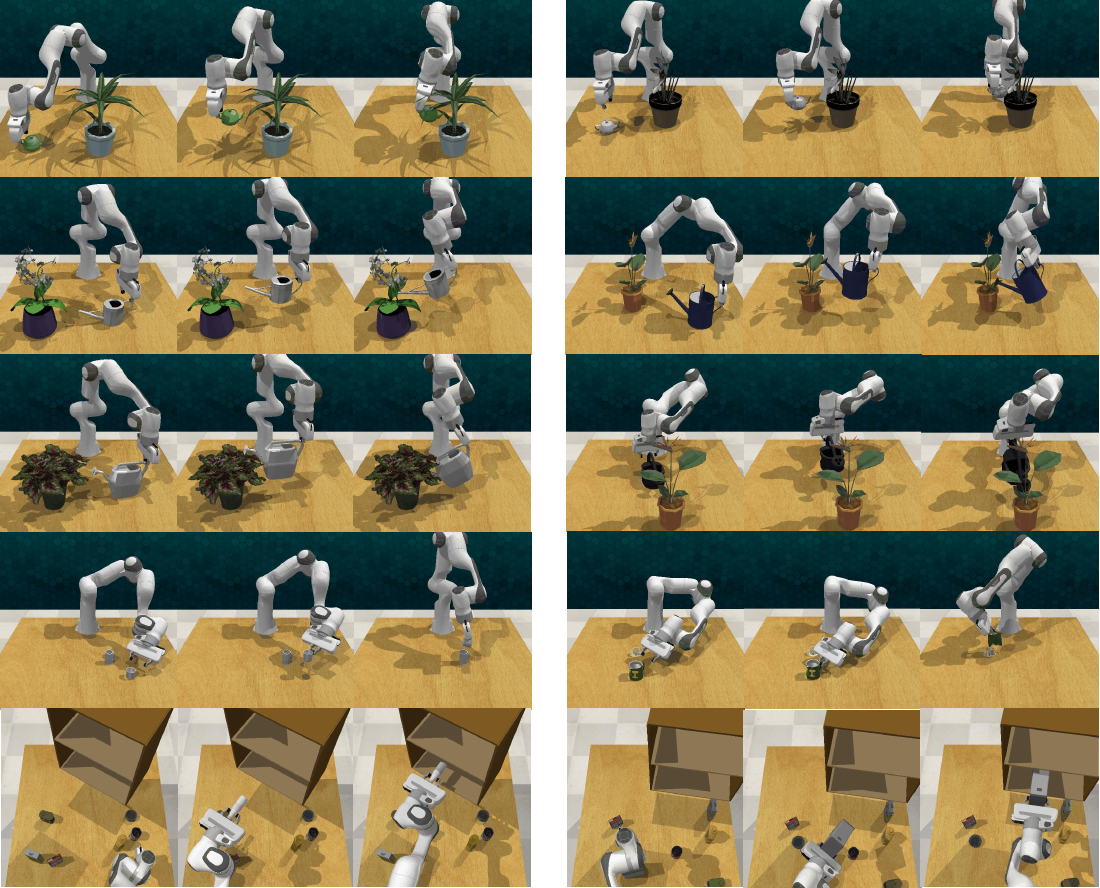}
    \caption{Additional 2D visualization results of automatic trajectory transfer in different instance-level and category-level variants.}
    \label{fig.appendix.results_trajectory_transfer.1}
\end{figure*}

Additional results of trajectory transfer are shown in Fig.~\ref{fig.appendix.results_trajectory_transfer.1}.

%% file: tabs/tab_instance_count_with_category.tex
\begin{table}[t]
\centering
\caption{Instance counts and categories.}
\begin{tabular}{lcc}
\hline
Instance & Count & Category \\
\hline
Mug      & 27  & Container \\
Pitcher  & 30  & Container \\
Teacup   & 11  & Container \\
Teapot   & 72 (discarded) & Container \\
Box      & 10  & Object \\
Apple    & 20  & Object \\
Can      & 20  & Object \\
Cookie   & 21  & Object \\
Cabinet  & 5   & Receptacle \\
\hline
\end{tabular}
\label{tab:instance_stats}
\end{table}